\documentclass[letterpaper, 10 pt, conference]{ieeeconf}  

\IEEEoverridecommandlockouts                              
\usepackage{graphics} 
\usepackage{epsfig} 
\usepackage{mathptmx} 
\usepackage{times} 
\usepackage{amsmath} 
\usepackage{amssymb}  

\usepackage{siunitx}
\usepackage{comment}
\usepackage{url}

\usepackage{color}

\title{\LARGE \bf
Near-optimal Deep Reinforcement Learning Policies from Data for Zone Temperature Control}

\author{Loris Di Natale, Bratislav Svetozarevic, Philipp Heer, and Colin N. Jones
\thanks{This research was supported by the Swiss National Science Foundation under NCCR Automation, grant agreement 51NF40\_180545.}
\thanks{L. Di Natale, B. Svetozarevic, P. Heer are with the 
Urban Energy Systems Laboratory, Swiss Federal Laboratories for Materials Science and Technology (Empa), 8600 Dübendorf, Switzerland {\tt\small\{loris.dinatale, bratislav.svetozarevic, philipp.heer\}@empa.ch.}}%
\thanks{L. Di Natale and C.N. Jones are with the 
Laboratoire d’Automatique, Swiss Federal Institute of Technology Lausanne (EPFL), 1015 Lausanne, Switzerland {\tt\small colin.jones@epfl.ch.}}%
}

\begin{document}






\maketitle
\thispagestyle{empty}
\pagestyle{empty}

\begin{abstract}

Replacing poorly performing existing controllers with smarter solutions will decrease the energy intensity of the building sector. Recently, controllers based on Deep Reinforcement Learning (DRL) have been shown to be more effective than conventional baselines. However, since the optimal solution is usually unknown, it is still unclear if DRL agents are attaining near-optimal performance in general or if there is still a large gap to bridge. 

In this paper, we investigate the performance of DRL agents compared to the theoretically optimal solution. To that end, we leverage Physically Consistent Neural Networks (PCNNs) as simulation environments, for which optimal control inputs are easy to compute. Furthermore, PCNNs solely rely on data to be trained, avoiding the difficult physics-based modeling phase, while retaining physical consistency. Our results hint that DRL agents not only clearly outperform conventional rule-based controllers, they furthermore attain near-optimal performance.


\end{abstract}

\section{Introduction}

In the European Union, buildings are responsible for $36\%$ of the energy-related greenhouse gas emissions, consuming $40\%$ of the end-use energy \cite{ec2019tfactsheet}. Almost two-thirds of the latter can be linked to space heating and cooling \cite{eurostat2020households}, which calls for solutions to decrease the energy intensity of the sector. 
Building occupants play a major role 
through the generated heat gains and the constraints that arise from their comfort requirements \cite{oldewurtel2013importance}. The latter transform the building control problem into a multi-objective optimization, where one tries to minimize energy consumption while maximizing the comfort of the users. To attain this objective, one solution is to replace poorly performing conventional zone temperature Rule-Based Controllers (RBCs) with more advanced methods, such as Deep Reinforcement Learning (DRL) policies \cite{svetozarevic2022data}. Indeed, numerous studies have already shown how DRL agents are able to improve upon the performance of classical RBCs in terms of energy consumption, cost savings, and/or comfort satisfaction, e.g. in \cite{zou2020towards, touzani2021controlling, brandi2022comparison, chen2021enforcing, valladares2019energy}. 

Because of the slow thermal dynamics of buildings and the data-inefficiency of DRL algorithms, months of data are routinely needed to converge to a good solution, e.g. in \cite{touzani2021controlling, nagy2018deep}. This is infeasible in real case studies and researchers thus have to rely on simulation environments. 
Many of them, e.g. \cite{touzani2021controlling, chen2021enforcing, valladares2019energy}, are trained in complex physics-based simulation environments, such as EnergyPlus \cite{crawley2001energyplus}. Setting up such environments can however be a very difficult and lengthy procedure \cite{zhang2019whole}, and it furthermore results in nonlinear models. 
On the other hand, only historical data is required to fit black-box models, such as in \cite{zou2020towards}, but one cannot ensure that such models are physically accurate and they remain highly nonlinear. Consequently, studies relying on such complex models usually do not provide upper bounds on the performance of DRL agents since the theoretical optimal solution (i.e. knowing the noise and disturbances) is often impossible to compute.

To the best of the authors' knowledge, works comparing DRL agents to the optimal performance rely on relatively simple models built from first principles, such as \cite{nagy2018deep, huang2022mixed}. In this paper, we consider predefined dynamic comfort bounds, which is an improvement over previously studied fixed bounds. Moreover, \cite{nagy2018deep} trains discrete DRL agents and \cite{huang2022mixed} assumes access to solar photovoltaic electricity generation and battery storage.
Furthermore, in this paper, we use Physically Consistent Neural Networks (PCNNs) \cite{di2021physically} as simulation environments instead of first principles models. PCNNs are trained from historical data, hence bypassing the complex design stage of physics-based methods, but remain physically consistent with respect to control inputs. However, they introduce nonlinearities in the model 
through the use of Neural Networks (NNs). Nonetheless, 
we can compute 
\textit{super-optimal} control inputs, i.e. 
assuming access to a perfect oracle of the future, with Linear Programs (LPs), which are easy to solve. This allows us to assess the performance of DRL policies in various settings in feasible time. 


To summarize our contribution, we analyze the performance of DRL agents for zone temperature control compared to industrial rule-based baselines and to the super-optimal solution, using PCNNs as simulation environments. Our pipeline solely relies on past historical data to fit the PCNNs, avoiding the difficult physics-based modeling procedure while retaining physical consistency. The agents are trained in continuous action spaces to maintain the temperature of the zone between predefined dynamic comfort bounds. We then investigate the sensitivity of the obtained policies to the choice of random seed and reward function, analyzing the gap with the theoretically optimal solutions in each setting. 

The rest of the paper is structured as follows. Section~\ref{sec:background} briefly introduces DRL, PCNNs, and the case study. We then detail our methods in Section~\ref{sec:methods}, present and discuss the results in Section~\ref{sec:results}, and conclude the paper with Section~\ref{sec:conclusion}.


\section{Background}
\label{sec:background}

\subsection{Deep Reinforcement Learning}

Reinforcement Learning (RL) problems are usually formulated as Markov Decision Processes (MDPs), which are represented by tuples $<S,A,P,\rho_0,r,\gamma>$, where $S$ is the state space, $A$ the action space, $P=P(s'|s,a)$ the probability of transitioning from state $s\in S$ to $s'\in S$ when action $a\in A$ is taken, $\rho_0$ the initial state distribution, $r=r(s,a)$ the reward function, 
and $0<\gamma<1$ the discount factor. At each time step $t$, the agent observes the state $s_t$ and chooses an action $a_t$. The environment then steps to the next state $s_{t+1}$, 
provides the reward $r(s_t,a_t)$ and $s_{t+1}$ to the agent, and the process repeats itself until the end of an episode after some time $H$.

The goal of RL algorithms is to find the policy $\pi(a|s)$ that maximizes the expected return $R_{t_0}$, i.e. the expected sum of discounted rewards, over an episode starting at time $t_0$:
\begin{align}
    J(\pi) = \mathbb{E}_{s_{t_0}\sim\rho_0, a\sim\pi}[R_{t_0}] = \mathbb{E}_{s_{t_0}\sim\rho_0, a\sim\pi}[\sum_{t=t_0}^{t_0+H}\gamma^{t-t_0}r(s_t, a_t)]
\end{align}
To that end, many algorithms rely on learning an approximation of this objective, named the Q-function, which estimates the expected return the agent will receive if it takes action $a$ in state $s$ and then follows the current policy $\pi$:
\begin{align}
    Q^\pi(s,a) = E_{a\sim\pi}[R_{t_0}|s_{t_0}=s, a_{t_0}=a]
\end{align}
The Q-function is often referred to as the critic, while the policy function is called the actor. In DRL, one parametrizes both the critic and the actor with NNs, i.e. $\pi_\theta(a|s)$ and $Q^\pi_\phi(s,a)$, and optimizes over $\{\theta, \phi\}$. 
Among the countless improvements and techniques presented in various contributions, we want to point out the influence of target networks \cite{van2016deep}, which are used in this work. The idea is to keep a copy of the actor and the critic in memory and only update these copies slowly to decrease the usual overestimation bias of Q-values and stabilize the learning process.

\subsection{Physically Consistent Neural Networks}

While classical NNs have been proven to be very effective function approximators in various fields, they remain agnostic to the underlying physical laws when modeling physical systems and might fail to grasp fundamental principles \cite{di2021physically}. For example, when modeling the temperature in a zone, turning the heating on should always have the effect of increasing the temperature, 
as DRL agents could otherwise learn spurious behaviors. To still leverage the expressiveness of NNs while retaining physical consistency with respect to some inputs, PCNNs were proposed and analyzed in \cite{di2021physically}, and we refer the reader to the original paper for the details. The main feature of PCNNs is that they are physically consistent with respect to power inputs, external temperatures, and temperatures in neighboring zones \textit{by design}. In particular, this means that applying heating or cooling power to the zone will have the expected impact on the change in temperature, ensuring that DRL agents learn meaningful behaviors.


In brief, the core idea of PCNNs is to treat part of the inputs -- the ones that require physical consistency -- in parallel of the main recurrent NN pipeline, in a linear module $E$. Heavily inspired from classical physics-based resistance-capacitance models to ensure consistency, this module  
accumulates or dissipates energy at each time step to capture the impact of the heating/cooling power $u$, the outside temperature $T^{out}$, and the temperature in the neighboring zone $T^{neigh}$ on the temperature of the modeled zone $T$. Simultaneously, a recursive NN pipeline $D$ processes the other inputs $x$ to capture unforced dynamics, i.e. the evolution of the temperature if no power is applied and heat losses are neglected. The final temperature prediction $T$ is then the sum of both components, the unforced dynamics and the accumulated energy since the beginning of the prediction. Altogether, we can write PCNN predictions as follows:
\begin{align}
    D_{k+1} &= D_k + f(D_k, x_k) \label{equ: nn} \\
    E_{k+1} &= E_k + a u_k - b (T_k - T^{out}_k) - c (T_k - T^{neigh}_k) \label{equ: energy accumulator} \\
    T_{k+1} &= D_{k+1} + E_{k+1}, \label{equ: T=D+E}
\end{align}
where $f$ is any nonlinear function, typically comprised of recurrent NNs, and $a$, $b$, $c$ are parameters learned simultaneously with $f$ through backpropagation. While this represents the situation during the heating season, the cooling case is analogous, using another parameter $d$ instead of $a$ to capture the efficiency of the cooling system instead of the heating one. Note that we define heating as positive and cooling as negative power throughout this paper, so that cooling decreases the energy accumulated in $E$, as expected.

\subsection{Case study}

\begin{figure} 
\centering
\includegraphics[width=\columnwidth]{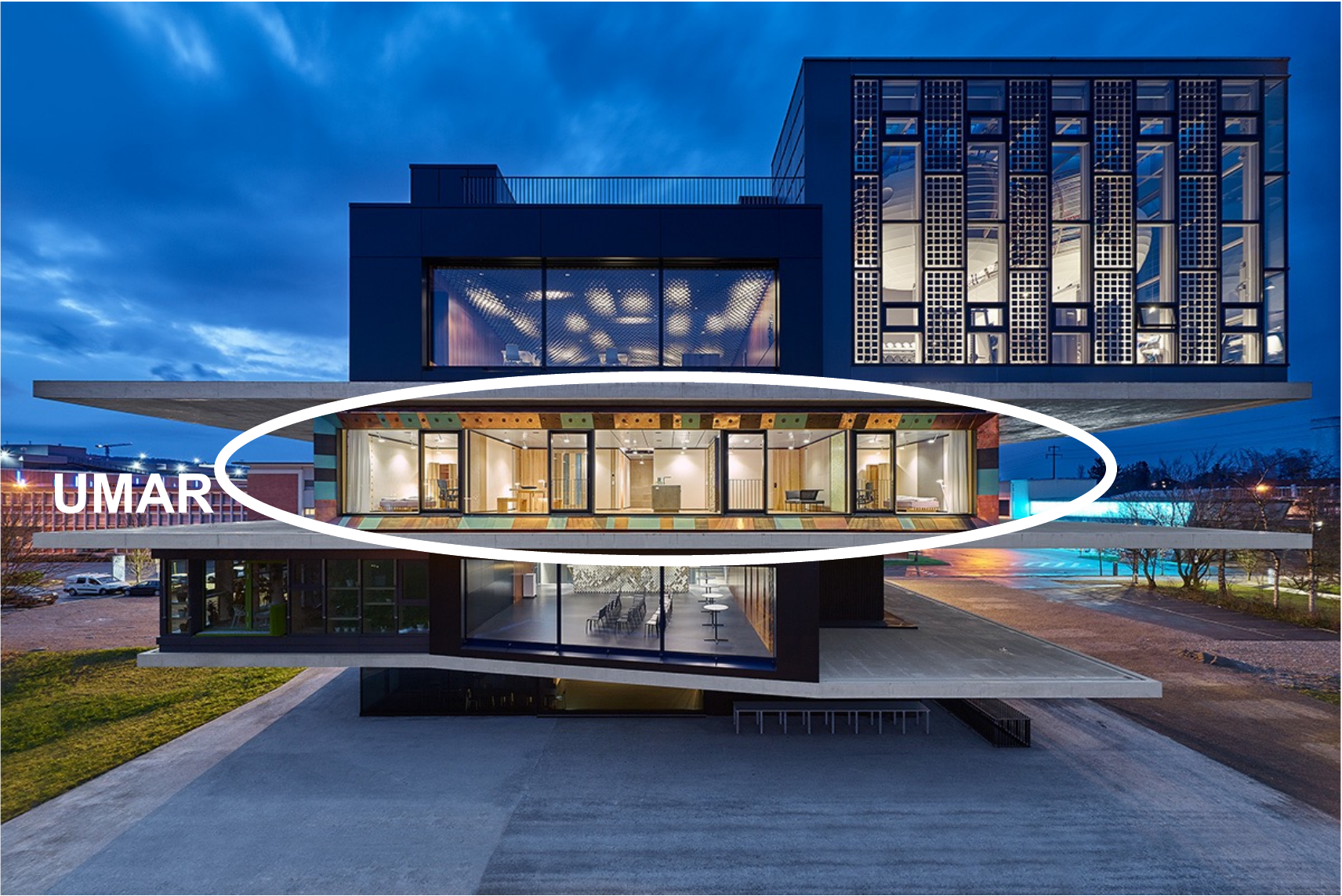}
\caption{NEST building, Duebendorf, and the UMAR unit circled in white {\copyright} Zooey Braun, Stuttgart.}
\label{fig:nest}
\end{figure}

The data used to train the PCNNs and DRL agents is taken from the Urban Mining and Recycling (UMAR) unit, an apartment in the NEST building, located in Duebendorf, Switzerland, and pictured in Fig.~\ref{fig:nest} \cite{nest}. It consists of two bedrooms separated by a living room, all powered by heating/cooling panels letting hot/cold water flow through their ceiling. More than three years of data is available and was processed as explained in \cite{di2021physically}, including the normalization procedure.

\section{Methods}
\label{sec:methods}

\subsection{Problem setting}

The task of the agents is to control the heating or cooling power of one bedroom in UMAR at each time step of \SI{15}{\minute}, i.e. $a_t = u_t$, to minimize the energy consumption while respecting predefined dynamic comfort bounds. While we require the temperature to stay between \SI{23}{\celsius} and \SI{24}{\celsius} during the night, i.e. from \SI{20}{\hour} to \SI{8}{\hour}, it can be relaxed during the day, when the bedroom is generally unoccupied. Specifically, we relax the lower bound to \SI{21}{\celsius} during the day in the heating season, and the upper bound is raised to \SI{26}{\celsius} from \SI{8}{\hour} to \SI{20}{\hour} during the cooling season.

The state $s$ observed by the agents at each time step is composed of the input features of the PCNN, i.e. zone temperatures (of the controlled and neighboring room), ambient conditions (temperature and solar irradiation), 
and time information (sine and cosine functions of the month of the year and time of the day, and the day of the week) \cite{di2021physically}. Additionally, agents know whether they can heat or cool and the current temperature comfort bounds for the occupants. To have more expressive policies aware of the evolution of the environment in time, we also add $12$ autoregressive terms, with \SI{15}{\minute} time steps, of the zone temperatures and ambient conditions, so that agents know the state of these variables during the last three hours when taking decisions. 



To train and evaluate the agents, we create \SI{12}{\hour} to \SI{72}{\hour}-long sequences of data with no missing values, hereafter referred to as \textit{trajectories}. These sequences, possibly overlapping each hour, are then separated into a training and a validation data set. 
To learn policies that are robust to measurement noise, we add a small independent Gaussian noise on top of the zone temperature measurement.

The reward function is defined as the negative weighted sum of comfort violations, i.e. how far from the designed bounds the temperature inside the zone is, and energy consumption. Mathematically, we thus have:
\begin{align}
    r(s_t,a_t) &= - \max{\{L_t-T_t, 0\}} - \max{\{T_t-U_t, 0\}} \nonumber \\
               &\ \quad - \lambda |a_t|, \label{equ:reward}
\end{align}
where $L$ and $U$ represent the lower and upper comfort bounds, respectively, 
and $\lambda$ the weighting factor. As a rule of thumb, we designed $\lambda$ such that the agent receives the same penalty for using a power of \SI{1}{\kilo\watt} and for being \SI{0.5}{\celsius} outside of the comfort bounds. 

\subsection{Algorithm}

In this work, we rely on the Twin Delayed Deep Deterministic policy gradient algorithm (TD3) \cite{fujimoto2018addressing}, an actor-critic method improving upon the classical Deep Deterministic Policy Gradient algorithm (DDPG) \cite{lillicrap2015continuous}. The main issue with DDPG is that the critic network has a tendency to sharply overestimate Q-values, which is then exploited by agents and can lead to poorly performing policies \cite{fujimoto2018addressing}. To correct this bias, TD3 introduces three main improvements over DDPG:
\begin{itemize}
    \item Inspired from the success of Double Q-learning \cite{hasselt2010double}, two critic networks are learned in parallel, and the smallest of the two approximated Q-values is used to limit overestimation.
    \item To avoid instability arising from fast-changing Q-functions, the actor and the targets networks are updated less frequently than the critics.
    \item To reduce the ability of the policy to exploit overestimated Q-values, noise is added to the action chosen by the actor before it is evaluated by the critics. 
\end{itemize}
In our implementations, all the NNs have three hidden layers of $512$ neurons and we use a slightly modified version of the Adam optimizer \cite{liu2019radam} with a learning rate of $10^{-4}$.

\subsection{Performance assessment}

To analyze the performance of the DRL agents, we compare them to two classical RBC baselines and the optimal solution. \textit{Baseline 1} is tracking a reference \SI{0.5}{\celsius} away from the bound, turning the heating on and off as soon as the target temperature is met. \textit{Baseline 2} is a classical rule-based controller with a one-degree hysteresis, i.e. it starts heating at full power when the temperature reaches the lower bound and until one degree has been gained. During the cooling season, similar strategies are used, with the controllers starting to cool once the upper bound or the reference temperature half a degree from it is reached.


To compute the super-optimal control inputs over an episode, knowing all the external conditions and the measurement noise $N$ over the horizon, we need to solve the following LP from time $t_0$, where the objective function is designed to match the reward of the agents:
\begin{align}
    \min_{u_0, ..., u_{H-1}}\quad  \sum_{k=0}^{H-1}&\tilde\lambda u_{k} + \epsilon^{L}_{k+1} + \epsilon^{U}_{k+1} \\
    s.t.\qquad \          
                              E_{0} &= E(t_0)\\
                              E_{k+1} &= E_k + a u_k - b (T_k - T^{out}_k) \nonumber \\
                              &\qquad \quad - c (T_k - T^{neigh}_k)\\
                              T_{k+1} &= D(t_{k+1}) + E_{k+1} + N(t_{k+1}) \label{equ: D external}\\
                              - T_{k+1} &\leq - L_{k+1} + \epsilon^{L}_{k+1} \\
                              T_{k+1} &\leq U_{k+1} + \epsilon^{U}_{k+1} \\
                              u^{low} &\leq u_k \leq u^{high},
\end{align}
where all the constraints hold for $k=0,...,H-1$. Note that we use $\tilde\lambda$ here, which equals $\lambda$ in heating cases and $-\lambda$ during the cooling season, so that it always penalizes the absolute value of the power used, as in the reward function \eqref{equ:reward}. The key property of PCNNs rendering this optimization feasible is the fact that the highly nonlinear unforced dynamics $D$ are independent of the control inputs $u$, which appear linearly in the temperature predictions of PCNNs in \eqref{equ: nn}-\eqref{equ: T=D+E}. $D$ can hence be computed \textit{a priori} for the entire horizon (knowing all external conditions) and fed into the optimization procedure as an external variable to compute the temperature evolution in \eqref{equ: D external}. This LP can be solved very efficiently with common tools, which allows us to compute the optimal solution for thousands of trajectories in feasible time. 
The code and data can be found here: \url{https://gitlab.nccr-automation.ch/loris.dinatale/NoDRL}.

\section{Results and Discussion}
\label{sec:results}

\subsection{Performance analysis}
\label{sec:analysis}

\begin{figure} 
\centering
\includegraphics[width=\columnwidth]{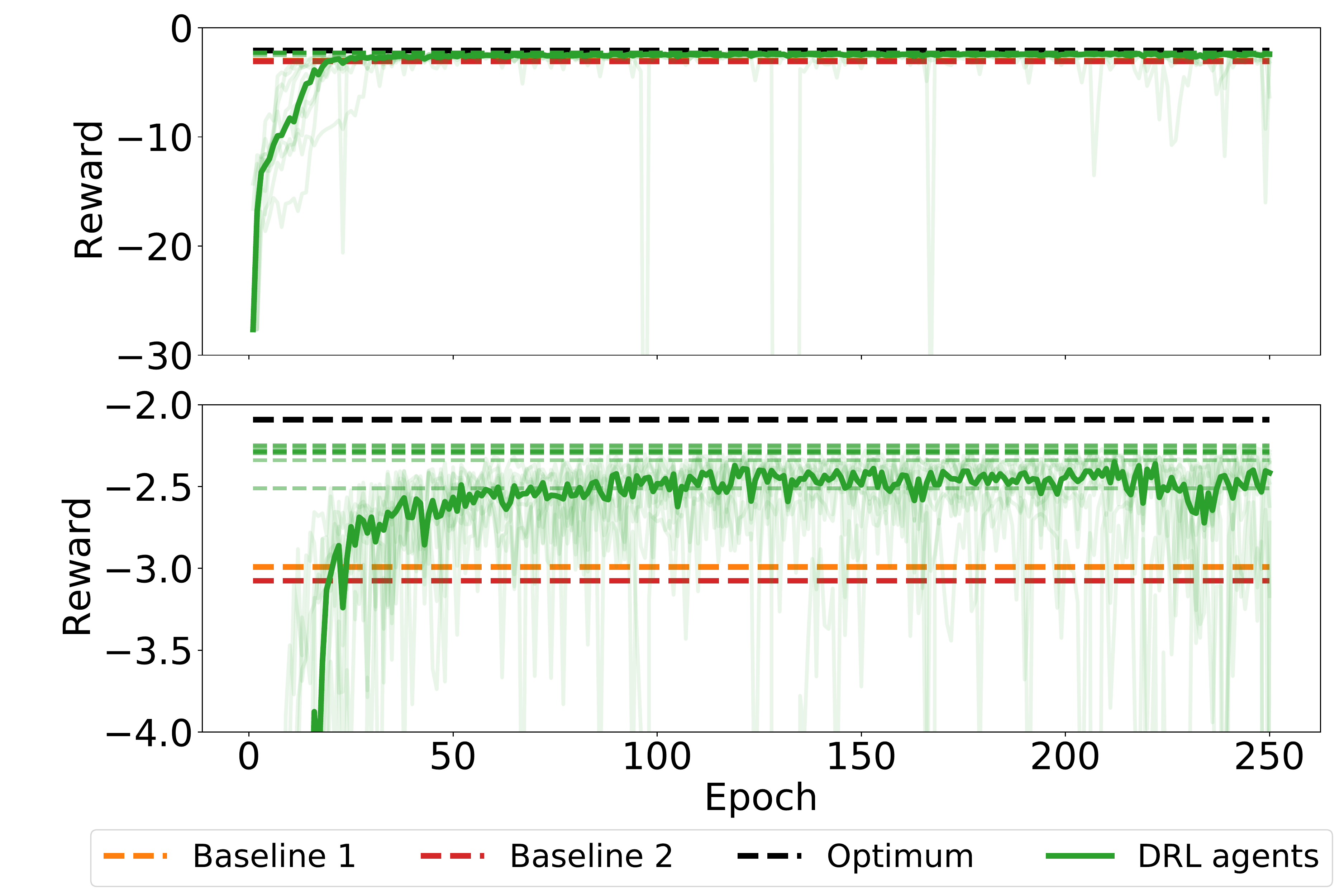}
\caption{Convergence rate of DRL agents with different random seeds in green, with the median in bold and the maximal reward attained by each agent in dashed lines. For reference, the baseline and optimal performance are also shown in dashed lines.}
\label{fig:convergence}
\end{figure}

Since DRL policies are notoriously sensitive to the random seed used \cite{henderson2018deep}, we trained $10$ agents with different seeds to compare their performance. In Fig.~\ref{fig:convergence}, we plot the convergence rate of the agents in green, with the median in bold, and zoom in the bottom plot to get a clearer picture. Here, one epoch represents $5000$ time steps in the environment, i.e. slightly over $50$ days of data, after which the agent is evaluated on $50$ trajectories from the validation set. The maximal performance attained by each agent of the training horizon is reported in dashed lines, and we included the baseline and optimal performance for reference. One can see the reward obtained by each agent fluctuating a lot, and most agents attained a maximal reward between $-2.25$ and $-2.3$. All DRL agents present comparable learning patterns: they converge to policies that perform similarly to the baselines after $35$-$40$ epochs and then consistently outperform them, with a decrease in the variance of the obtained rewards. However, we can see a few exceptions along the training pattern, with some agents' performance plummetting for a few epochs but recovering quickly. 
It is also noteworthy that one of the agents performed significantly worse than the other ones, only reaching a best reward of $-2.51$. This proves that DRL agents are also sensitive to the choice of random seed in our setting, and one should hence always perform several runs to rule out the possibility of the random seed impacting the quality of the results.

\begin{figure}
\centering
\includegraphics[width=\columnwidth]{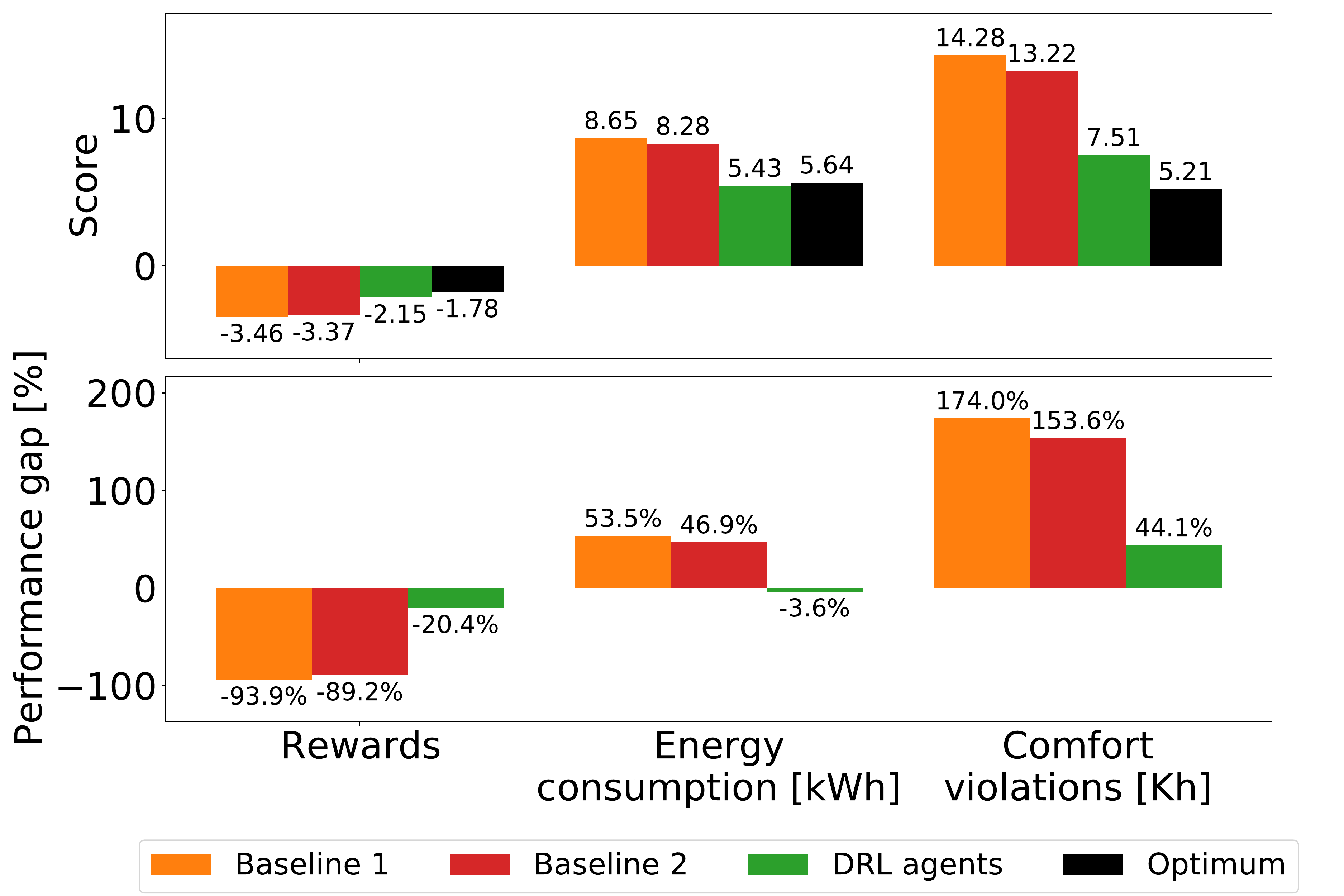}
\caption{Comparison between the baselines, the average performance of three DRL agents, and the optimal solution on the validation set. The top plot presents the average performance of each controller after subtraction of the unavoidable penalties, while the bottom one show the gap in performance with respect to the optimal trajectories for each controller.}
\label{fig:barplot}
\end{figure}

\begin{figure} 
\centering
\includegraphics[width=\columnwidth]{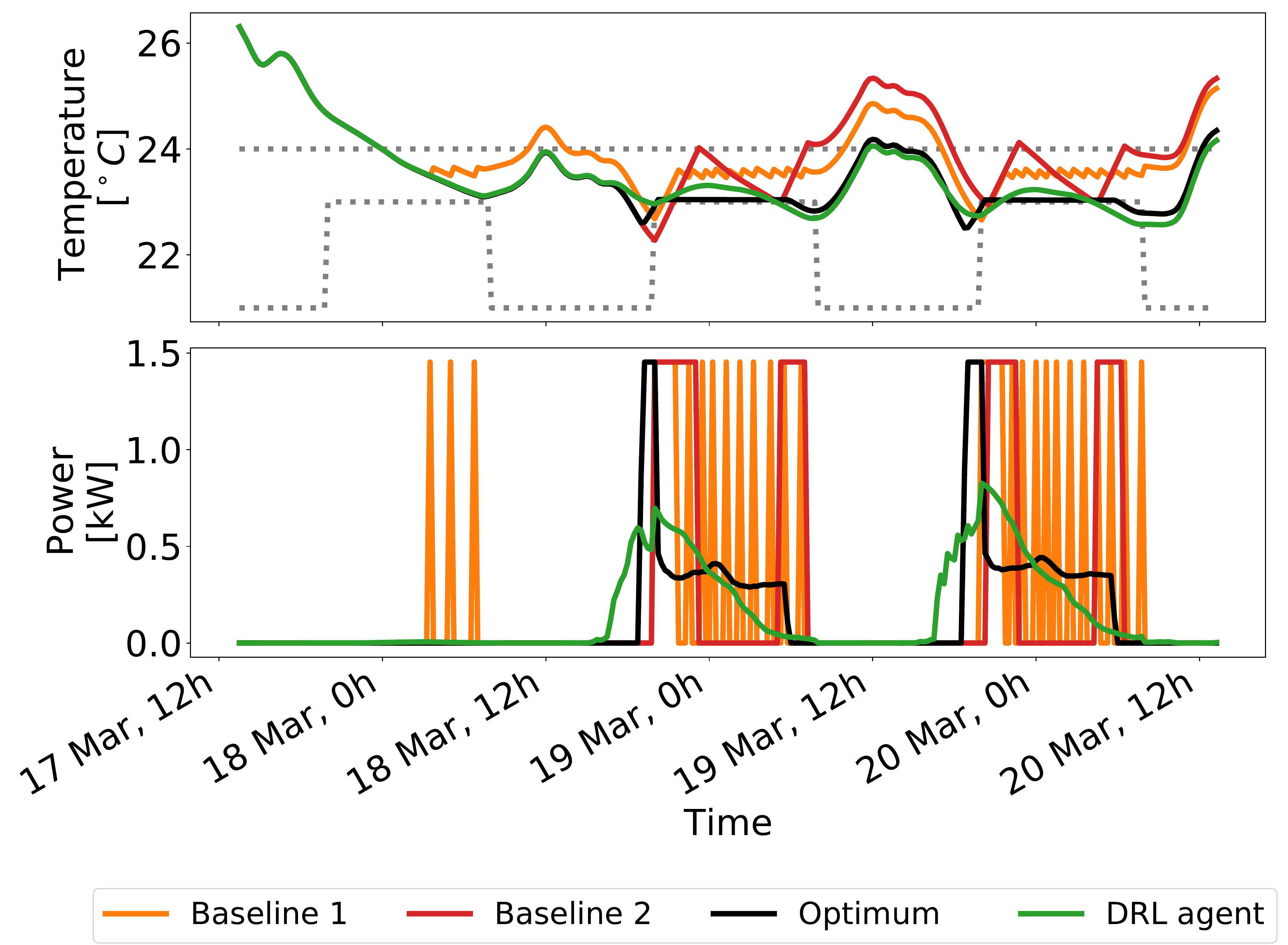}
\caption{Example of the behavior of a DRL agent over $3$ days, compared to the baselines and the optimal solution. The top plot shows the room temperature, the bottom one the power input.}
\label{fig:timeseries}
\end{figure}

For each agent, we selected the best policy obtained during the training phase, i.e. the one that achieved the maximum reward on the $50$ test trajectories, and analyzed its average performance on almost $2000$ $3$-day long trajectories from the validation set in terms of rewards, energy consumption, and comfort violations. In the top plot of Fig.~\ref{fig:barplot}, one can observe the median results achieved by the different random seeds, where this metric was preferred over the mean to decrease the impact of the poorly performing seed. Note that the random seed also impacts the performance of the baselines and the optimal solution because it changes the noise added to the temperature measurements. However, very little variance could be observed between different runs, the influence of the Gaussian noise cancels out over the entire validation data set and was hence not reported. The bottom plot shows the performance gap between the optimal one and the other controllers. Note that we subtracted \textit{unavoidable} penalties from all statistics: since each episode is initialized from a situation observed in the historical data, the temperature in the zone might be initialized out of the defined comfort bounds, leading to unavoidable penalties for any controller (see Fig.~\ref{fig:timeseries}). We removed them to give a better picture of the difference between each controller. 

Remarkably, with the chosen parameters, agents were able to converge to a near-optimal solution. They found a relatively similar trade-off between energy consumption and comfort violations as the optimal solution, 
consuming roughly the same amount of energy as the optimal solution at the cost of slightly increased comfort violations of a little over $20\%$. This is, however, still much better than what the baselines are able to do, both in terms of energy savings and comfort improvements. 
Alltogether, these results confirm that DRL agents are able to converge to policies that not only clearly beat classical controllers but actually attain near-optimal performance. 

In general, all agents were able to capture the expected and desired preheating/precooling behavior. Indeed, they always take action earlier than the other controllers, especially the rule-based ones, to anticipate constraint tightenings, as for example pictured in Fig.~\ref{fig:timeseries} (without Gaussian noise for clarity). They usually heat or cool the zone until the temperature is slightly above or below the comfort bound and then stop. This often results in the temperature reaching the bound again just before the constraints are relaxed. Overall, this strategy is not far from the optimal one, which starts to act later before the constraints tighten and uses the full available power to reach the narrow comfort bounds faster. It then takes advantage of its full knowledge of the environment to input just enough energy in the system for the temperature to stay exactly at the desired limit and avoid comfort penalties unless, for example, 
solar heat gains are expected to create more violations later (e.g. Fig.~\ref{fig:timeseries}). Due to their reactive nature, both baselines cannot anticipate constraint tightenings and relaxations, acting too fast or too late. Furthermore, they do not anticipate any external input, such as the solar irradiation, for example leading to overheating behavior.

In the particular case of Fig.~\ref{fig:timeseries}, the agent consumed slightly less energy than the optimal solution, \SI{10.29}{\kilo\watt\hour} against \SI{10.66}{\kilo\watt\hour}, but had slightly more comfort violations, \SI{13.2}{\kelvin\hour} against \SI{12.4}{\kelvin\hour}. The difference can be noticed when the comfort bounds were relaxed: the agent stopped heating slightly earlier to save energy and anticipate potential solar gains, while the optimal solution kept heating to stay longer above the lower bound. In general, DRL agents have a tendency to converge to risk-averse polices because of the noisy observations returned by the environment. Indeed, they usually slightly overheat/overcool the zone when the constraints are tightened to end up further away from the bounds and hence avoid comfort penalties arising from the noisy temperature measurement jumping outside of the comfort zone. On the other hand, when computing the optimal trajectory, the noise is assumed to be known, such that the optimization knows when to input power to the zone to always stay in bounds despite the impact of the noise.

\addtolength{\textheight}{-0cm}   

\subsection{Sensitivity to the weighting factor}

\begin{figure} 
\centering
\includegraphics[width=\columnwidth]{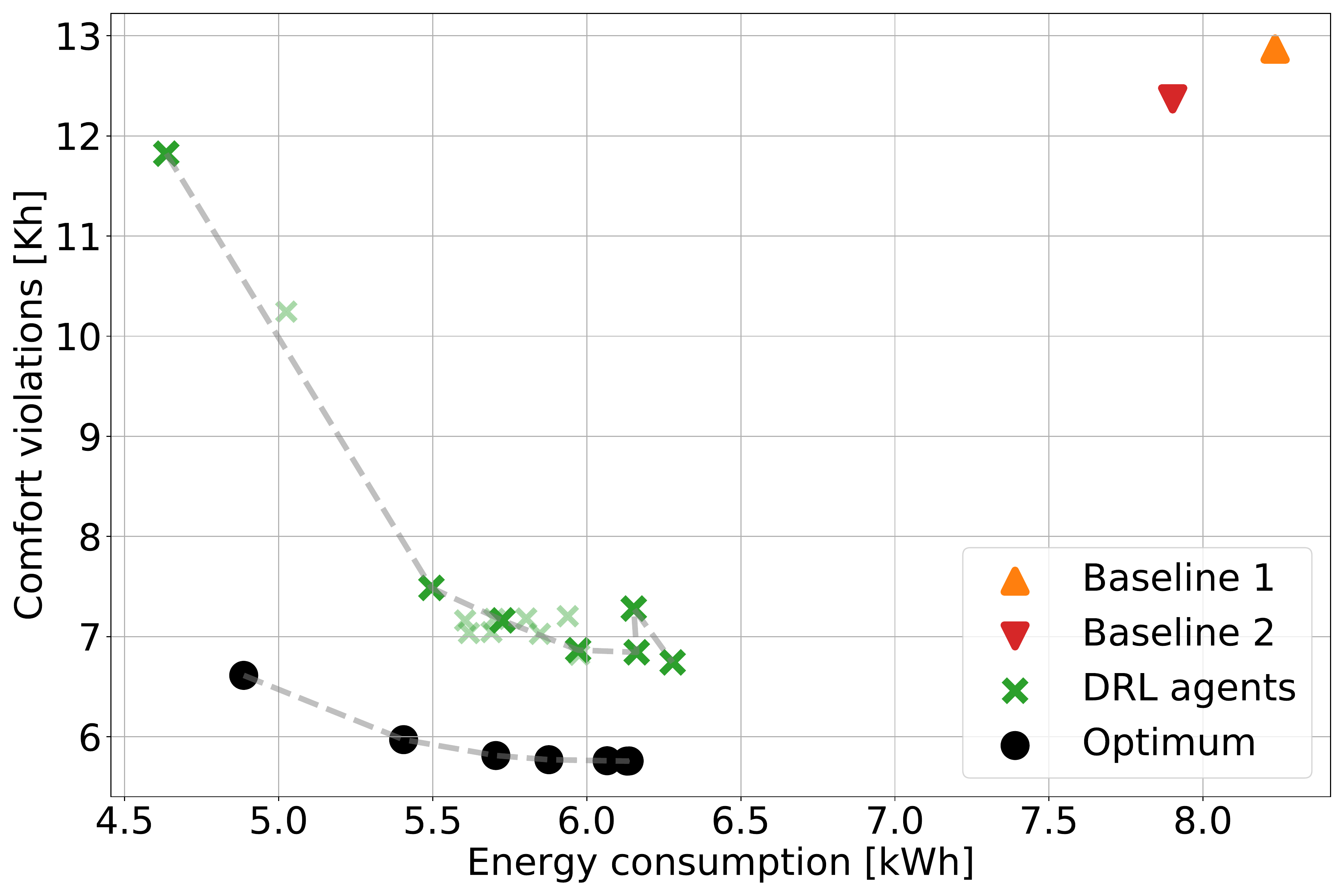}
\caption{Sensitivity analysis of DRL agents and the optimal solution to different weighting factors in \eqref{equ:reward}, decreasing it from left to right along the dashed lines. The trade-off obtained by both baselines is also plotted for reference, as well as the results obtained with different seeds in Section~\ref{sec:analysis} in shaded green.}
\label{fig:sensitivity}
\end{figure}

To assess the impact of the weighting factor $\lambda$ on the policy found by the DRL agents, we performed a sensitivity analysis to investigate the range of values that lead to near-optimal solutions and the different trade-offs reached by the different policies. To that end, we trained agents in the same environment, with a fixed random seed, but multiplying or dividing $\lambda$ by increasing powers of two to reflect situations where more and more importance is put on decreasing the energy consumption or the amount of comfort violations, respectively. We can then analyze the Pareto frontiers arising from the trade-off between both objectives for the DRL policies and the optimal solutions in  Fig.~\ref{fig:sensitivity}, where the weighting factor was decreased from $4\lambda$ to $\frac{1}{16}\lambda$ from the left to the right along the dashed Pareto frontiers. The performance of the baselines was also plotted for reference.

As can be observed, the agent is generally able to strike a trade-off similar to the optimal one, usually consuming roughly the same amount of energy at the cost of slightly more comfort violations, as long as the weighting factor is not too big. Once $\lambda$ is multiplied by $4$, the agent struggles to find an interesting solution: it uses very little energy but does not improve the comfort of the occupants much compared to the baselines (top left green point). When we tried to increase $\lambda$ by a factor of $8$, the agent quickly converged to a very poor policy that never uses energy at all. On the other hand, decreasing the weighting factor impacts the quality of the solution less: policies consume slightly more energy each time, slowly reducing the amount of comfort violations. 

In this analysis, it is important to remember that the choices of random seed again impacts the results, which could explain why the trade-off obtained by the agent with a weighting factor of $\frac{1}{8}\lambda$ is slightly higher than the Pareto front. This is confirmed by the shaded green markers showing the performance of the agents discussed in Section~\ref{sec:analysis}, where we see the impact of different random seeds for the same weighting factor. All random seeds usually lead to similar solutions, but we can also observe the outlier pointed out previously (top left shaded green marker). Interestingly, even though it obtained worse rewards than all the other agents, we can see its behavior still lying near the Pareto front. Hence, it would seem that this agent simply converged to a different solution that was not optimal in this situation but might be the expected behavior under different circumstances. Note that the choice of random seed also impacts the performance of the baselines and the optimal solution, but we again found that the differences were small and hence only plotted one solution for clarity.

To summarize, one should be careful with the design of the weighting factor, as it might impact the quality of the solution. However, our tests showed that a wide range of values, from $2\lambda$ to $\frac{1}{16}\lambda$ in our case, could be selected, reflecting different preferences for the occupants, and still lead to near-optimal behaviors. In general, choosing a value that is too large seems to be more problematic than the contrary. Finally, while the random seed seems to impact in which region of the Pareto frontier the solution converges, our experiments provide evidence that DRL agents are always converging to a region near the Pareto frontier of all DRL policies. However, the latter is diverging from the optimal Pareto frontier when agents try to use too little energy, and it thus seems safer to choose small values for $\lambda$.

\section{Conclusion}
\label{sec:conclusion}

In this work, leveraging the expressiveness and physical consistency of PCNNs, we trained DRL agents to control the temperature of a building zone, balancing energy consumption and comfort violations. Thanks to the structure of PCNNs, we were able to compute the theoretical optimal control inputs over each episode, which showed that DRL agents were not only clearly outperforming rule-based baselines, they were also able to obtain near-optimal policies. We analyzed the impact of the random seed and of the weighting factor balancing energy consumption and comfort violations on the obtained solutions. The former was shown to sometimes lead to lower quality solutions in terms of the obtained rewards, but still lying near the Pareto frontier of both objectives. For the latter, we found a large range of possible values leading to near-optimal solutions, until the factor is too big.

Interestingly, the entire pipeline to design and fit PCNNs and DRL agents only relies on historical data, avoiding the difficult physics-based design phase. However, 
DRL agents still take years of artificial data to converge to interesting policies. In future works, we thus want to address this data-inefficiency issue and create faster converging DRL agents through physics-inspired structures. We also plan to scale up our analysis to larger and more complex systems, and to compare DRL agents to Model Predictive Controllers, both in simulation but also on real buildings.

\section{Acknowledgments}

The authors gratefully acknowledge the support of the Swiss National Science Foundation.


\bibliographystyle{IEEEtran}
\bibliography{biblio.bib}



\end{document}